\pdfoutput=1

\documentclass[11pt]{article}

\usepackage[]{EACL2023}

\usepackage{times}
\usepackage{latexsym}

\usepackage[T1]{fontenc}

\usepackage[utf8]{inputenc}

\usepackage{microtype}

\usepackage{inconsolata}
\usepackage{graphicx}
\usepackage{multirow}
\usepackage{hyperref}
\usepackage{booktabs}
\usepackage[normalem]{ulem}
\usepackage[framemethod=TikZ]{mdframed}

\usepackage{xspace}
\usepackage{xcolor}

\title{Personalized Text Generation with Fine-Grained Linguistic Control}

\newif\ifreviewer
\reviewertrue

\ifreviewer

\fi

\author{Bashar Alhafni\textsuperscript{\textnormal{1}}, Vivek Kulkarni\textsuperscript{\textnormal{2}}, Dhruv Kumar\textsuperscript{\textnormal{2}}, Vipul Raheja\textsuperscript{\textnormal{2}} \\
  \textsuperscript{1}New York University Abu Dhabi\\
  \textsuperscript{2}Grammarly\\
  \texttt{alhafni@nyu.edu},   \texttt{firstname.lastname@grammarly.com}
  }

\definecolor{raspberry}{HTML}{de6767}
\definecolor{blue(x11)}{HTML}{4689f2}

\begin{document}
\maketitle
\begin{abstract}

As the text generation capabilities of large language models become increasingly prominent, recent studies have focused on controlling particular aspects of the generated text to make it more personalized. However, most research on controllable text generation focuses on controlling the content or modeling specific high-level/coarse-grained attributes that reflect authors' writing styles, such as formality, domain, or sentiment. In this paper, we focus on controlling fine-grained attributes spanning multiple linguistic dimensions, such as lexical and syntactic attributes. We introduce a novel benchmark to train generative models and evaluate their ability to generate personalized text based on multiple fine-grained linguistic attributes. We systematically investigate the performance of various large language models on our benchmark and draw insights from the factors that impact their performance. We make our code, data, and pretrained models publicly available.\footnote{\url{https://github.com/balhafni/personalized-gen}} 


\end{abstract}

\section{Introduction}

\begin{figure}[t]
    \centering
     \includegraphics[width=\linewidth]{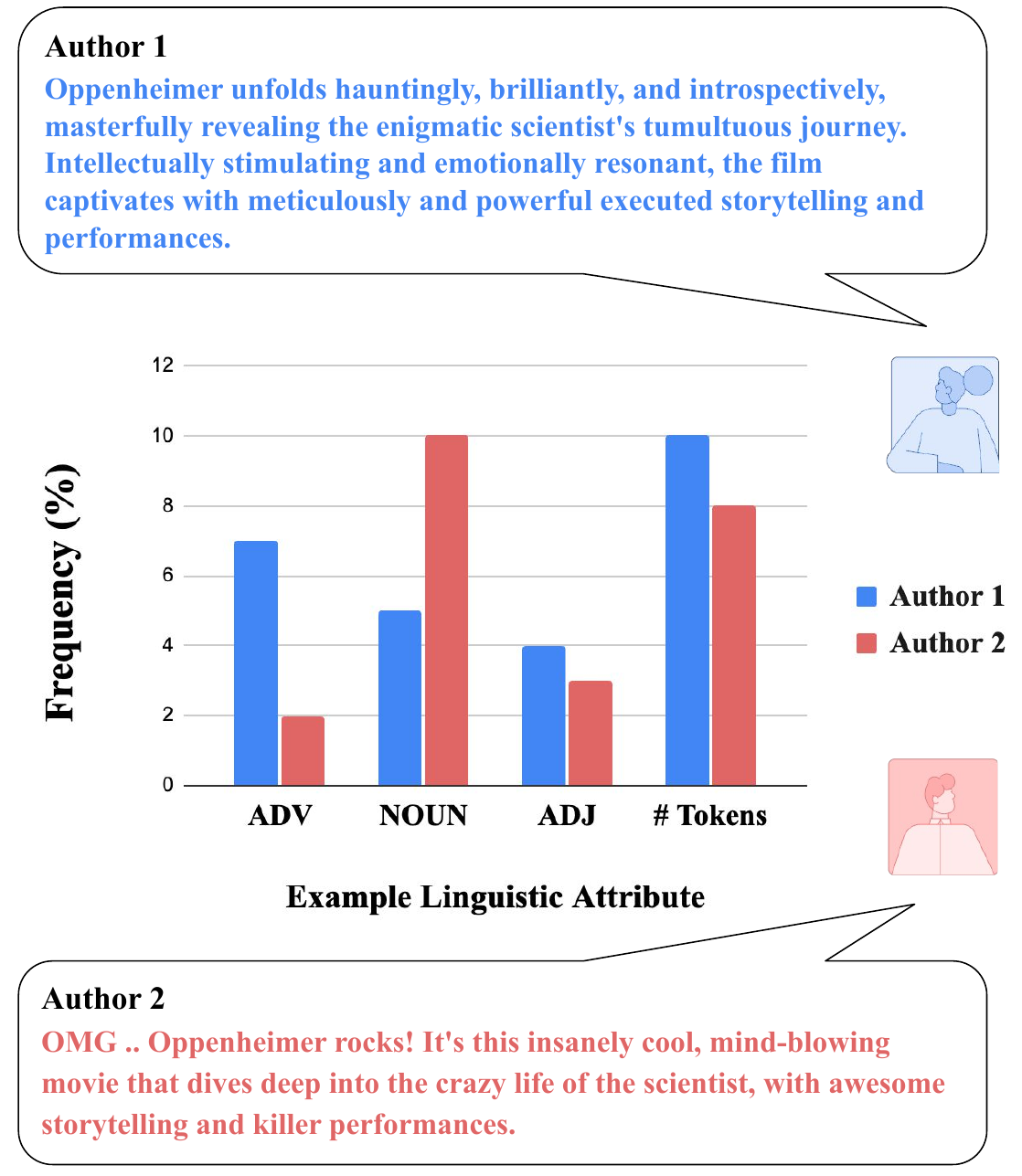}
    \caption{An example highlighting the stylistic linguistic differences between two reviews about the \href{https://en.wikipedia.org/wiki/Oppenheimer_(film)}{Oppenheimer} movie. The bar chart indicates the frequencies of the adverbs (ADV), nouns (NOUN), adjectives (ADJ), and the number of tokens (\# Tokens). The \textcolor{blue(x11)}{blue} and \textcolor{raspberry}{red} bars represent the frequencies based on the reviews written by \textcolor{blue(x11)}{Author 1} and \textcolor{raspberry}{Author 2}, respectively.}
\label{fig:example}
\end{figure}

With the evolution of large language models (LLMs), applications involving text generation have become increasingly prominent \cite{kaddour2023challenges}. In recent years, there has been a flurry of writing and conversational assistants that harness the power of LLMs at each stage of the writing life-cycle~\cite{raheja-etal-2023-coedit,gomez-rodriguez-williams-2023-confederacy}. At the same time, there has been a significant emphasis on research that enables the generation of personalized text -- text that reflects the author's style and content more accurately. For instance, Figure~\ref{fig:example} presents two movie reviews that are conveyed in different ways. The reviews differ linguistically in terms of the number of tokens and the distributions of adverbs, adjectives, and nouns. 


Several works explore the personalization of various NLP tasks such as Lexical Simplification \cite{lee-yeung-2018-personalizing, bingel-etal-2018-lexi}, Dialogue Modeling \cite{zhang-etal-2018-personalizing, mazare-etal-2018-training, madotto-etal-2019-personalizing, 10.1145/3404835.3462828, zhong-etal-2022-less}, Machine Translation \cite{rabinovich-etal-2017-personalized}, Grammatical Error Correction \cite{nadejde-tetreault-2019-personalizing}, Summarization \cite{takatsu-etal-2021-personalized}, and Title Generation \cite{salemi2023lamp} among others.
However, a key question remains unanswered: How effective are LLMs at personalizing the text they generate? While undoubtedly, there has been a lot of work on controllable text generation (e.g., using either control codes \cite{keskarCTRL2019} or adapters \cite{hu2023adaptertst}), many of these works model the author's style either by using a unique control code or a dense embedding and do not really evaluate how effective these models are at generating text that reflects a fine-grained set of stylistic features. For instance, authors may exhibit variations in the distribution of discourse markers in their writing.
To what extent can models reflect fine-grained stylistic differences and account for them when generating text? We seek to advance this line of research further by introducing a novel benchmark that evaluates the ability of generative models to accurately reflect fine-grained stylistic attributes spanning multiple linguistic dimensions.
Our benchmark enables us to gain valuable insights into how model-generated text responds to deliberate variations in style across different dimensions, providing a thorough understanding of the model's sensitivity.
In addition to developing a benchmark, we systematically explore and analyze the performance of LLMs on this benchmark and draw key insights into the efficacy of various models and what factors drive performance. Moreover, we also propose improvements to baseline models that can significantly improve performance on this newly introduced benchmark for multi-attribute personalized style transfer.  In a nutshell, our key contributions are as follows:
\begin{itemize}
\item \textbf{Benchmark} We introduce a benchmark for the task of personalized text generation with fine-grained linguistic control.
\item \textbf{Linguistic Feature Analysis} We perform an extensive characterization of the authors' styles in terms of linguistic features spanning across multiple linguistic dimensions including lexical, rhetoric, and syntactic features.
\item  \textbf{Improved Models} We evaluate the effectiveness of various baseline models on this task and provide insights on the performance of current LLMs.

\end{itemize}

\begin{table}[]
    \centering
    \begin{tabular}{lcc}
    \toprule
         \textbf{Dataset} & \textbf{\# Docs} & \textbf{\# Authors}\\\hline
         \textbf{Blogs} & 24,913 & 140 \\
         \textbf{IMDb62} & 38,693 & 62 \\
         \textbf{Amazon} & 42,542 & 49 \\\hline
         & 106,148 & 251 \\\hline
        
    \end{tabular}
    \caption{Data statistics of our benchmark.}
    \label{tab:data-summary}
\end{table}

\section{Benchmark Construction}
In this section, we discuss the construction process of our benchmark in terms of datasets, linguistic attributes, and evaluation metrics. 

\subsection{Task Definition}
\label{sec:task-def}
Given a set of stylistic attributes encapsulating the writing style of a particular author and an input textual prompt, the goal is to generate text that adheres to the provided attributes. More formally, we define a set of $n$ attributes $\{a_1, a_2, ..., a_n\}$, where each attribute $a_i$ has a set of possible values $\{v_1, v_2, ... v_k\}$. Given a possible assignment to these values, our goal is to generate text that conforms to these values.

\subsection{Datasets}
\label{sec:datasets}
To construct our benchmark, we draw on multiple sources of author-grounded data spanning multiple domains. We chose samples from various datasets while ensuring that the number of authors we intend to model remains reasonable with respect to the models' capacity and the need to avoid excessive sparsity. We derive our benchmark from the following datasets:

\paragraph{Blogs Authorship Corpus} consists of 681.3K blogs written by 19,320 authors on \textit{blogger.com} in 2004 \cite{schler2006}. The dataset includes information about the authors in terms of their age, gender, industry, and astrological signs. Each author is associated with one of 40 industrial categories. We select the largest five industrial categories in terms of the number of blogs to include in our benchmark. Namely, we focus on the following categories: Technology, Education, Arts, Internet, and Communication-media. Out of this subset, we select blogs that have at least 50 words and authors who wrote at least 100 blogs each. In the end, we ended up with 24,913 blogs written by 140 authors.

\paragraph{IMDb62} contains 62,000 movie reviews written by 62 prolific users of the Internet Movie Database (IMDb) where each user wrote 1,000 reviews \cite{seroussi-etal-2014-authorship}. Out of all reviews, we only keep the ones with at least 50 words. In total, we ended up with 38,693 reviews written by 62 authors.

\paragraph{Amazon Reviews} comprises 233.1M product reviews on Amazon published between 1996 and 2018 \cite{ni-etal-2019-justifying}. We use the publicly available 5-core version of the corpus which consists of 157.3M reviews written by 9.8M users where each user wrote at least five reviews. Out of this rich set, we selected users who wrote at least 2,800 reviews with each review containing at least 50 words. This resulted in 42,542 reviews written by 49 authors.

Table~\ref{tab:data-summary} presents a summary of the datasets we use in our benchmark. Putting all the selected samples together leaves us with 106,148 textual examples written by 251 authors to constitute our benchmark. For each author, we randomly sample examples to create train (80\%), development (10\%), and test (10\%) splits. This ensures that all 251 authors modeled during training will also have examples in the development and test sets. Altogether, we have a total of 84,824 examples for training (Train), 10,506 examples for development (Dev), and 10,818 examples for testing (Test).






\begin{table}[t]
    \centering
    \begin{tabular}{ll}
    \toprule
        \textbf{Features Type}  & \textbf{Attribute}  \\\hline 
        \multirow{3}{*}{Lexical} & \# Tokens  \\
                & \# Sentences  \\
                & Readability Score \\\hline
         \multirow{2}{*}{Morpho-Syntax} & \# POS Tags \\
                      & \# Syntax Relations  \\\hline
        Discourse & \# Rhetorical Relations  \\
        \bottomrule

    \end{tabular}
    \caption{Summary of the linguistic attributes we model.}
    \label{tab:attributes}
\end{table}

\subsection{Controllable Linguistic Attributes} 
\label{sec:attributes}
To operationalize an author's linguistic style, we extract various linguistic attributes from the author's text. In particular, we represent each author's linguistic style as a vector of attributes -- linguistic attributes that capture patterns in lexical usage, morpho-syntactic information, and discourse markers. More specifically, for each author \textbf{A}, we associate a feature vector $\phi(A)$ that represents the author's linguistic style. Each dimension of $\phi(A)$ corresponds to attributes from the following feature families as summarized in Table~\ref{tab:attributes}:

\paragraph{Lexical Features} We model the lexical usage for each author by using three lexical features: \textit{number of tokens}, \textit{number of sentences}, and the {readability} level for each text written by a specific author. We use spaCy \cite{spacy2} to tokenize and parse the text to obtain the number of tokens and the number of sentences. We measure the \textit{readability} of the text (i.e., the ease with which a reader can understand the text) using the FKGL score \cite{Flesch1948}. 

\paragraph{Morpho-Syntactic Features} We consider the frequencies of the part-of-speech (POS) tags and the dependency relations to capture the morpho-syntactic information for each author's text. For the POS tags, we consider 14 core tags as defined in the Universal POS tagset, which are internally subdivided into open-class words (i.e., adjective, adverb, interjection, noun, proper noun, verb), closed-class words (i.e., adposition, auxiliary, coordinating conjunction, determiner, numeral, particle, pronoun, subordinating conjunction), and the class of `other' which includes punctuation and symbols. For the dependency relations, we use 32 dependency relations as defined by ClearNLP.\footnote{\url{https://github.com/clir/clearnlp-guidelines}} We use the POS tagger and the dependency parser available in spaCy to extract the morpho-syntactic information. 

\paragraph{Discourse Features}  Discourse coherence, which refers to how well a text or speech is organized to convey information, captures important stylistic aspects of one's writings. Rhetorical Structure Theory (RST) \cite{mann:1988} is one of the most influential approaches for document-level discourse analysis. It can represent a document’s discourse structure using a hierarchical tree in which nodes are recursively linked with rhetorical relations and labeled with nucleus or satellite tags to depict the importance of the child nodes in a relation. In our work, we consider the frequency of the RST relations for each author's text to capture discourse coherence. We use the publicly available RST parser\footnote{\url{https://github.com/EducationalTestingService/rstfinder}} introduced by \newcite{heilman2015fast} to obtain RST relations. We model 3 RST relations in total.


Once we extract the above linguistic features for each textual example written by a particular author, we average the values of the features for all examples to obtain the author's vector. We consider the above features since they succinctly capture the author's linguistic style in an interpretable manner (i.e., each vector dimension has a clear interpretation), and further, such vector representation serves as a lever for controlled modification of specific linguistic attributes. Table \ref{tab:atts-vals} in Appendix~\ref{app:atts-list} provides the complete list of all the attributes we consider for this benchmark.

\subsection{Linguistic Attributes Representation}
\label{sec:disc-process}

After extracting the linguistic attributes, each attribute will be represented by a continuous value indicating the average frequency of that attribute in texts written by a specific author. To reduce sparsity and enhance the models' abilities to learn contextual representations for each attribute value, we discretize the values for all attributes. To do so, we group the values for each attribute into deciles based on the training data. After that, each value is represented by a specific discrete bin corresponding to the range it falls within. Representing each attribute using discrete bins reduces the vocabulary size the model has to learn to reflect the specified linguistic attributes in the generated text. 


\subsection{Evaluation Metrics} 
\label{sec:eval_metrics}
We evaluate models on their effectiveness of generating (or rewriting) input text that corresponds to a given style (as operationalized by the attributes discussed in \S\ref{sec:attributes}).
To quantify the effectiveness of various models at responding correctly to controlled changes in style, we report the success rate as measured by the fraction of times the model generated text whose style vector corresponds to the specific style provided. We compute the success rate for each attribute as well as the relative improvement over a random baseline (i.e., choosing an attribute bin at random). We report the \textbf{mean success rate} and the \textbf{median of the relative improvements} across all attributes as a summary of model performance. Naturally, the mean success rate ranges from 0 to 100. However, the relative improvements depend on the number of bins for each attribute; generally, it ranges from -100 to (\# of bins) $\times$ 100 for each attribute. Moreover, we use the distribution of grammatical errors to compare the \textbf{fluency} of the generated text against the gold data for each author. We use an in-house grammatical error detection model to get the distributions of grammatical errors. For each author, we compare the error distributions by running a t-test \cite{yuen:1974} and report the fraction of authors whose distributions are not statistically significant ($p > 0.05$). Hence, the fluency scores will range from 0 to 100.  




\section{Models}

\begin{figure*}[t]
    \centering
    \frame{\includegraphics[width=\linewidth]{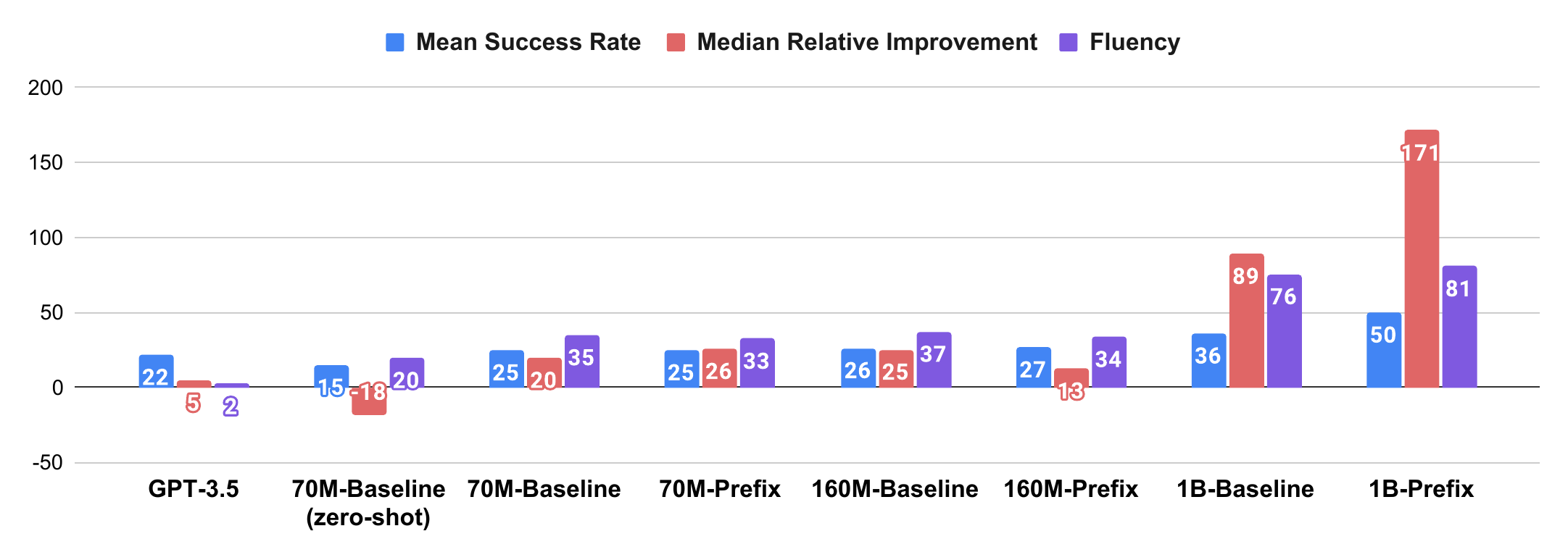}}
    \caption{Results on the Test set of our benchmark using GPT-3.5 and the 70M, 160M, and 1B Pythia models. The Mean Success Rate and Fluency vary from 0 to 100. The Median Relative Improvement can vary from -100 to (\# of bins) $\times$ 100 for each attribute. Details on the evaluation metrics are provided in  \S\ref{sec:eval_metrics}.
    }
\label{fig:dev-results}
\end{figure*}

To evaluate our proposed benchmark and given the ubiquitous adoption of causal language models (CLMs) for various NLP tasks recently, we consider the models from the \textbf{Pythia} Scaling Suite \cite{biderman2023pythia}. The suite comprises eight models with sizes ranging from 70M to 1B parameters. All the models are based on the GPT-Neo architecture \cite{gpt-neo}, which is an open-source replication of GPT-3 \cite{brown2020language}. The models were pretrained using the Pile dataset \cite{gao2020pile}, an 825GB English dataset containing texts from 22 diverse sources, roughly broken down into five categories: academic writing, internet, prose, dialogue, and miscellaneous. All the Pythia models were pretrained on the same data. In this work, we use the 70M, 160M, and 1B Pythia models.

\subsection{Baselines}
For our baselines, we fine-tune the Pythia models on our benchmark dataset without feeding any attributes to the models. During inference, we feed the first sentence of every textual example in the Dev or Test sets to get the generated predictions. The intuition behind this baseline is to investigate if the models are able to pick up on the linguistic attributes from the text itself. We also consider GPT-3.5 (ChatGPT)\footnote{\url{https://openai.com/blog/chatgpt}}, given the good performance it has shown on various NLP benchmarks.  We use the \texttt{gpt-3.5-turbo0613} model from the OpenAI API.\footnote{\url{https://api.openai.com}} We describe the training settings and the model's hyperparameters in Appendix~\ref{app:experiments}. The prompts we use for the GPT-3.5 experiments are also presented in the same appendix. 

\subsection{Multi-Attribute Controlled Models}
Given an input sentence along with an author linguistic vector, the goal of the models is to generate text that reflects the provided linguistic attributes (\S\ref{sec:task-def}). To incorporate features context into the models during training, we treat each author's linguistic attribute vector as a sequence of tokens and prepend it to the textual examples written by that particular author. We extend the input embedding layer of all models by adding the attributes and their values as additional tokens to the models. This enables the model to learn the semantics of the attributes. During inference, we prepend the author's linguistic attribute vector to the first sentence of every textual example in the Dev or Test sets to get the generated predictions. We refer to these models as \textbf{Prefix} throughout the paper.

\section{Experiments and Results}
In this section, we present the results of various models along with an investigation into the model's sensitivity to the attributes. We also explore how the training sample size influences the model's performance.

\subsection{Overall Results}
The results of the Test set are presented in Figure~\ref{fig:dev-results}. 

\paragraph{Baselines} Although GPT-3.5 achieves a mean success rate of 22, it did not do well in terms of the median relative improvement and fluency. It is worth noting that GPT-3.5's low score in fluency reflects the difference between the distribution of grammatical errors in the output of GPT-3.5 and the gold data. A sample of GPT-3.5 outputs reveals that the generated text is acceptable but contains different error proportions compared to the gold data. When it comes to the Pythia models, the untrained 70M Pythia baseline model (zero-shot) has the lowest scores in terms of the mean success rate and the median relative improvement. We observe an interesting pattern for the trained baselines: the larger the models, the better the results are across all metrics, with the 1B Pythia model being the best performer among the baselines. 

\paragraph{Multi-Attribute Controlled Models} The 70M and 1B Pythia \textit{Prefix} models achieve a better performance than their baseline counterparts. This highlights that, generally, the learned representations of the linguistic attributes provide the models with better control.
Overall, the 1B Pythia Prefix model was the best performer across all metrics. In terms of attribute-specific performance, we present the results of the 1B baseline and the 1B Prefix models in Appendix~\ref{app:attributes-results}. Across almost all attributes, the Prefix model was the better performer in terms of the success rate and relative improvements. 




\begin{figure}[t]
    \centering
    \includegraphics[width=\linewidth, trim={2cm 0 0 2cm}]{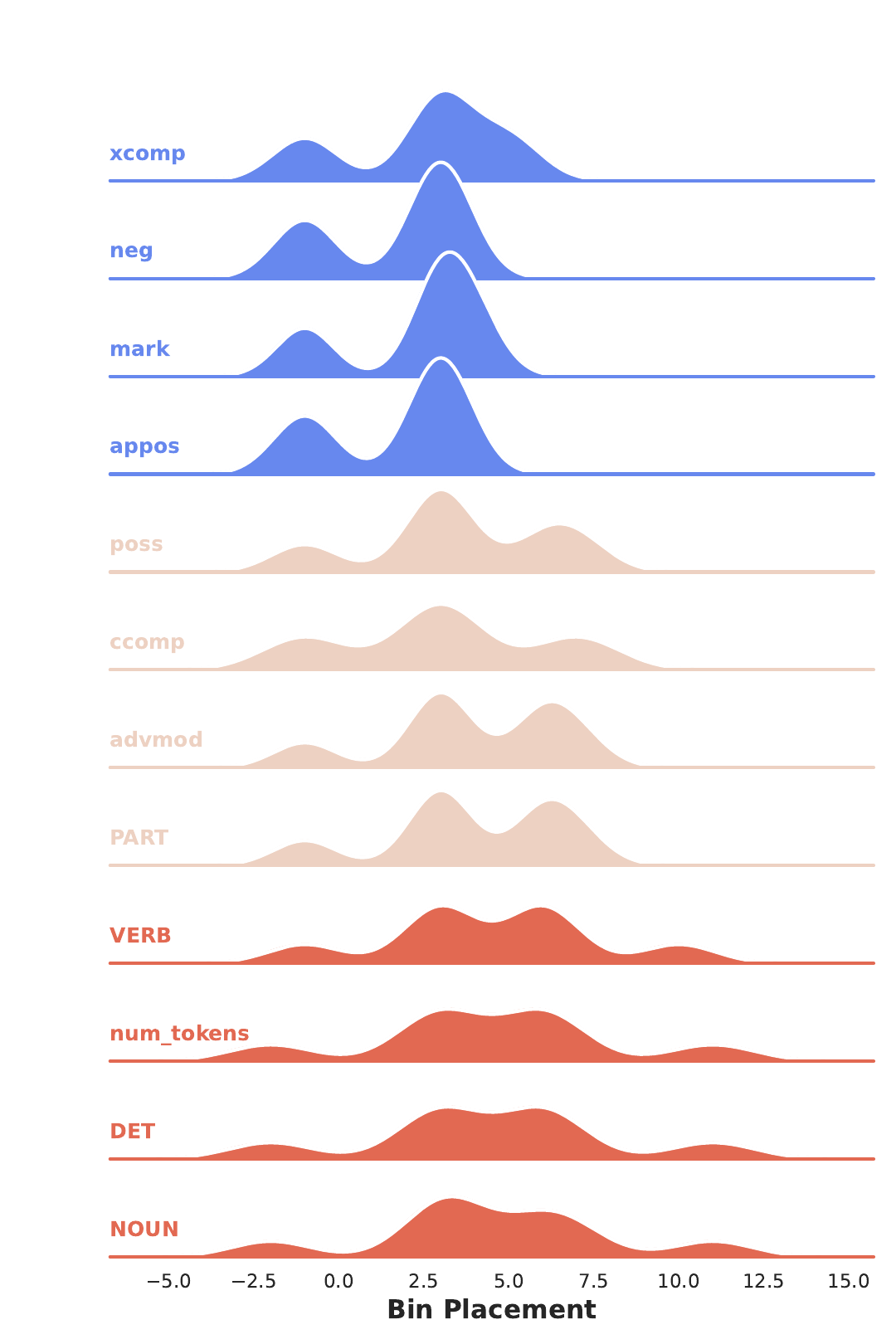}
    \caption{A collection of three different clusters of attributes, varying by their Bin Placement (x-axis) and their Mean Success Rate (\%, y-axis). Bin Placement refers to the number of bins skipped when assigning a new bin to each attribute (in sorted order), while the Mean Success Rate is calculated as explained in \S\ref{sec:eval_metrics}.
    We observe that the attributes display three distinct sensitivity patterns. 
    }
\label{fig:att-analysis}
\end{figure}

\subsection{Attribute Sensitivity Analysis}
We investigate the model's sensitivity to the attributes it has been conditioned on. To do so, we sample examples randomly from the Dev set for each author with at least 1000 data examples in Train. Then, for each attribute in each example, we change the attribute's respective bin according to the possible bins it could have while keeping all other attributes the same. This is done in a controlled setup where each attribute would be assigned to a range of possible bins that reflect an increase or a decrease from its gold value. During inference, we prepend the updated author's attribute vector to the first sentence of the textual example written by that particular author and perform greedy decoding to generate the model outputs. It is worth noting that we do this generation step for each attribute change. This resulted in 14,206 generated examples.

To measure the sensitivity of the model to the changes in the attributes, we compute the attribute-specific success rate by checking if the attribute value in the generated text moved in the right direction in terms of an increase or a decrease based on its newly assigned bin. We report the average success rate for each attribute across all generated examples. Figure \ref{fig:att-analysis} shows model performance as a function of attribute value displacement for 12 representative attributes.  From the figure, we can make the following two observations: (a) First, we observe that the attributes display three distinct sensitivity patterns. 
(b) Second, performance responses tend to be bi-modal or even tri-modal, with performance peaking at certain values and degrading as one falls off the extremes. We believe a deeper investigation that could explain the emergence of the observed multi-modal performance is an interesting one, which we leave to future work.

\begin{figure}[t]
    \centering
    \includegraphics[width=\linewidth]{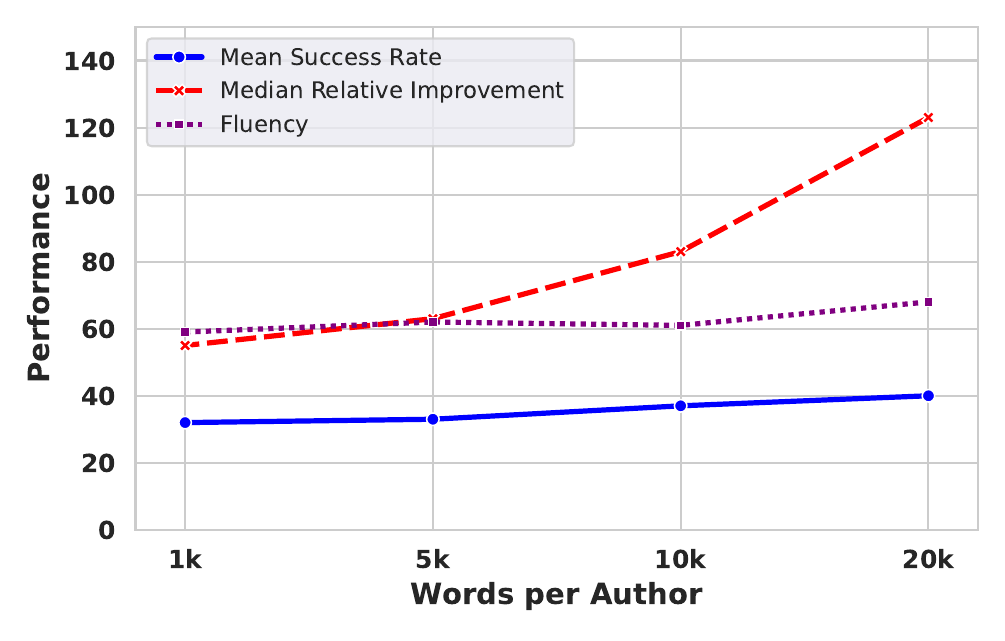}
    \caption{The performance of the 1B Pythia Prefix model on the Dev set at different data sizes in terms of the mean success rate, the median relative improvement, and fluency.}
\label{fig:learning-curve}
\end{figure}

Table~\ref{tab:gen-examples} presents examples of model-generated outputs. The examples illustrate how the outputs of different models vary due to changing specific attributes for a fixed input sentence, highlighting the models' attribute sensitivity.

\begin{table*}[!t]
  \centering
  \small
  \begin{tabular}{@{}l|p{0.2\textwidth}|p{0.65\textwidth}@{}}
    \toprule
    \textbf{Model} & \textbf{Attributes} & \textbf{Output} \\
    \midrule
    Original & \texttt{VERB}: 8-11, \newline \texttt{num\_sents}: 5-6, \newline ... & \textcolor{purple}{Today's lunch \underline{\textbf{was}} a \underline{\textbf{layered}} entree, \underline{\textbf{consisting}} of, from bottom to top, corn chips, \underline{\textbf{shredded}} lettuce and popcorn chicken.} What \underline{\textbf{was}} this culinary delight? A popcorn chicken salad, of course. I \underline{\textbf{have}} yet to \underline{\textbf{figure}} out exactly why there \underline{\textbf{were}} corn chips \underline{\textbf{involved}}. And why the salad bar then \underline{\textbf{had}} taco-plate fixin's, rather than salad fixin's.\\
    \midrule\midrule
    GPT-3.5 & \texttt{VERB}: 8-11, \newline \texttt{num\_sents}: 5-6, \newline  ... & \textcolor{purple}{Today's lunch \underline{\textbf{was}} a \underline{\textbf{layered}} entree, \underline{\textbf{consisting}} of, from bottom to top, corn chips, \underline{\textbf{shredded}} lettuce and popcorn chicken.} The dish \underline{\textbf{was}} a perfect combination of crunchy and savory flavors, with the corn chips \underline{\textbf{providing}} a satisfying crunch and the popcorn chicken \underline{\textbf{adding}} a deliciously salty taste. The shredded lettuce \underline{\textbf{added}} a refreshing touch to the dish, \underline{\textbf{making}} it a perfect meal for a hot summer day. Overall, the dish \underline{\textbf{was}} a great choice for lunch and I would definitely \underline{\textbf{recommend}} it to anyone \underline{\textbf{looking}} for a tasty and satisfying meal.
    \\    
    \midrule\midrule
    & \textbf{\texttt{VERB}: 27-32}, \newline \texttt{num\_sents}: 5-6, \newline ... & \textcolor{purple}{Today's lunch \underline{\textbf{was}} a \underline{\textbf{layered}} entree, \underline{\textbf{consisting}} of, from bottom to top, corn chips, \underline{\textbf{shredded}} lettuce and popcorn chicken.} The chicken \underline{\textbf{was}} \underline{\textbf{seasoned}} and \underline{\textbf{layered}} with chips. I \underline{\textbf{had}} one bag of popcorn chicken and the other \underline{\textbf{had}} regular chicken breast. It \underline{\textbf{was}} quite tasty and guilt-free. Of course, there \underline{\textbf{was}} the inevitable over-\underline{\textbf{eating}} comment from my hubby about how I don't \underline{\textbf{like}} eating chicken when there's so much stuff in it now. He \underline{\textbf{said}} he wouldn't \underline{\textbf{eat}} another chicken entree today. Shh, he's \underline{\textbf{talking}} shop. (He's not wrong!) \\\cmidrule{2-3}
    
    1B-Prefix  & {\texttt{VERB}: 8-11, \newline \textbf{\texttt{num\_sents}: >=21}, \newline ...} & \textcolor{purple}{Today's lunch \underline{\textbf{was}} a \underline{\textbf{layered}} entree, \underline{\textbf{consisting}} of, from bottom to top, corn chips, \underline{\textbf{shredded}} lettuce and popcorn chicken.} The chicken, which I've \underline{\textbf{had}} in the past, \underline{\textbf{was}} really great -- very moist and flavorful. There \underline{\textbf{were}} also some other good flavors \underline{\textbf{going}} on, but the entrees \underline{\textbf{were}} strong enough to \underline{\textbf{keep}} me from \underline{\textbf{feeling}} overwhelmed. I \underline{\textbf{had}} about a tablespoon of corn and a cup of pop corn, and it \underline{\textbf{was}} just right. Both flavors \underline{\textbf{were}} good, though the lettuce \underline{\textbf{was}} overbearing and I \underline{\textbf{was}} having a hard time \underline{\textbf{telling}} what \underline{\textbf{was}} real and what wasn't. Sharky's \underline{\textbf{has}} \underline{\textbf{convinced}} me that it's not always the food I \underline{\textbf{know}} in my mouth. That there's more to be \underline{\textbf{learned}} about what we \underline{\textbf{eat}}.\\

\\
    \bottomrule
\end{tabular}

\caption{Illustrative examples of how the generated outputs of different models vary due to changing specific attributes for a fixed input sentence. We present generated outputs using GPT-3.5 and the 1B Pythia Prefix model (our best model). Text in \textcolor{purple}{purple} indicates the input prompt. Attributes in bold (i.e., VERB or num\_sents) indicate a change in the bin for this attribute. Underlined and bold words refer to verbs.}

\label{tab:gen-examples}

\end{table*}

\subsection{Training Sample Efficacy}
 

To gain insights into the scaling behavior of personalized LLMs in terms of data, we examine the impact of the number of training samples per author on model performance by limiting the training data per author to 1k, 5k, 10k, and 20k words. It is worth noting that we repeat the discretization process described in \S\ref{sec:disc-process} for all attributes based on the scaled training data before training and evaluating the models. Figure~\ref{fig:learning-curve} presents the performance of the 1B Pythia Prefix model on the Dev set after being trained on different data sizes. As expected, we observe performance decrease across all three metrics as we reduce the number of words per author. We see the most degradation in performance in the median relative improvement. This is because it reports the median of the relative improvement to a random baseline and, therefore, has the highest sensitivity to the model performance. We also observe a degradation in the mean success rate; however, it is gradual as this metric is dependent on the bins, which makes it less sensitive to change. For fluency, we initially see a slight decrease in performance as we decrease the words per author to 10k, followed by a slight decrease for lower data proportions. For all three metrics, we see the biggest decrease in performance as we lower the training instances per author beyond 20k words.

\section{Related Work}

\subsection{Personalized Language Modeling}
Numerous works have explored learning user embeddings as a way to capture syntactic and semantic properties of the language of individuals \cite{hovy-2015-demographic, ouldamer:ujm-01377080, ijcai2017p547, welch-etal-2020-exploring, welch-etal-2020-compositional, rocca-yarkoni-2022-language}. These are then typically used to provide personalized inductive priors to downstream tasks. Recent studies \cite{mireshghallah-etal-2022-useridentifier, zhong-etal-2021-useradapter, oba2023perplm} have explored leveraging techniques such as adapters for adapting LLMs for various personalized NLP and NLG tasks.

Our work is closely related to \citet{king-cook-2020-evaluating}, who analyzed various methods, such as interpolation, fine-tuning, and priming language models for the personalization of general-purpose language models. They also analyzed model adaptation for models trained on users with similar demographics. This was inspired by \citet{lynn-etal-2017-human}, who showed that demographic factors could help model a variety of classification tasks, and found that personalized models perform better than those adapted from similar demographics. \citet{shao-etal-2020-examination} have also explored language models for personalization but focused on handling OOV tokens. 

Finally, our work is also related to \citet{salemi2023lamp} in that they also propose a benchmark for training and evaluating language models for personalized text generation tasks. However, their work is focused on specific tasks such as headline and email subject generation, whereas our work is focused on personalized language modeling. 

\subsection{Multi-Attribute Controlled Generation}
Controlling text generation using control codes is an active area of research where, given a code $c$, the learning problem is formulated to generate text $x$ by calculating the probability of $p(x|c)$. Numerous works have been proposed to train attribute-conditional models by fine-tuning pretrained models with attribute-specific corpora or training conditional generative networks \cite{keskarCTRL2019, Dathathri2020Plug, lample2018multipleattribute, 10.5555/3327345.3327417, krause-etal-2021-gedi-generative, russo-etal-2020-control, yu-etal-2021-attribute-alignment, kulkarni2021lmsoc}. Some of the noteworthy lines of research have proposed techniques such as Prefix-Tuning \cite{qian-etal-2022-controllable}, Adapters \cite{pmlr-v97-houlsby19a}, and Prompt-tuning \cite{yang-etal-2023-tailor, chen-etal-2023-mixture} to approach the problem. However, most of the prior works are limited by the number of attributes, as well as the amount of variation in those attributes that they are able to model, typically focusing on high-level or coarse-grained attributes such as sentiment or domain.

A major advantage of our approach is its ability to introduce a large number of discrete or continuous attributes, with a fair amount of variation in each of them. It focuses on a much greater number of fine-grained linguistic attributes that can be easily inferred from large amounts of data, irrespective of document lengths, domains, or other higher-order characteristics, which makes it better adaptable to a diverse range of applications. This adaptability is particularly beneficial in scenarios where the data varies significantly in style, structure, or content. Our approach leverages the inherent capabilities of pretrained models, enhancing them with attribute-specific features. This enables the model to not only understand but also generate text that accurately reflects the desired attributes, be they structural, syntactic, or stylistic. Moreover, our work allows for the dynamic adjustment of attributes in real-time, offering a level of control and precision that is not typically feasible with traditional models. As a result, users can tailor the output to meet specific requirements without the need for extensive retraining or manual intervention. This flexibility is highly essential and desirable for personalized text generation.


\section{Conclusion and Future Work}
We presented a novel benchmark to train and evaluate the ability of LLMs to generate personalized text based on multiple linguistic attributes. While existing research predominantly concentrates on content control or modeling specific writing style elements, our work stands out by focusing on nuanced stylistic attributes across diverse linguistic dimensions. We systematically investigate the performance of various LLMs and draw insights from the factors that drive such 
performance. We make our code, data, and pretrained models publicly available to encourage research on personalization and controlling LLMs. In future work, we plan to explore the use of other pretrained LLMs, and to extend the linguistic features that are covered in our benchmark.



\section*{Limitations}
In this work, while we present a benchmark to study the various linguistic phenomena in text, our study is limited to English. It would be interesting to observe how the patterns we observe for English extend to other languages. Although we investigate the performance of LLMs, we restrict ourselves to CLMs (GPT-3.5 and Pythia) for both the zero-shot and fine-tuning experiments. It would be interesting to study if this effect is replicated in other models, such as encoder-decoder-based models. We propose three automated metrics for our benchmark and observe interesting findings on models' capabilities to handle the fine-grained attributes. However, even though we do manually look at generated examples to ensure that our metrics make sense, a large-scale human study towards looking at generated outputs would help us study and better understand the current models' performance and capabilities. Finally, while building our dataset, we draw on multiple sources of author-grounded data that spans multiple domains; we could further extend this to more domains and authors.

\section*{Acknowledgments}
We thank our colleagues at Grammarly for the helpful discussions and constructive feedback.

\bibliography{anthology,custom}
\bibliographystyle{acl_natbib}

\appendix
\clearpage

\section{Experimental Setup}
\label{app:experiments}
\paragraph{Hyperparameters} We use Hugging Face's transformers~\cite{wolf-etal-2020-transformers} to fine-tune our models. We fine-tune the Pythia models for 10 epochs on 8 GPUs by using a learning rate of 5e-5, a batch size of 4 with 4 gradient accumulation steps, a seed of 42, and a maximum sequence length of 1024.  At the end of the fine-tuning, we pick the best checkpoint based on the performance of the Dev set.

\paragraph{GPT-3.5}
For the GPT-3.5 experiments, we used the prompt described in Figure \ref{fig:gpt_prompt}.

\begin{figure}[h]
\mdfdefinestyle{vipul}{innertopmargin=10pt,innerbottommargin=10pt,roundcorner=5pt}
\begin{mdframed}[style=vipul]
\small

Complete the given input sentence so that the stylometric attributes of the completed text are close to the provided stylometric attributes. The length of the auto-completed text should be about 1024 tokens.

\texttt{\textless stylometric vector\textgreater~}Attributes~\texttt{\textless/stylometric vector\textgreater} 
\texttt{\textless input\textgreater}~Input Text~\texttt{\textless/input\textgreater}

\end{mdframed}
\caption{Prompt template we used for the GPT-3.5 experiments.}
\label{fig:gpt_prompt}
\end{figure}

\onecolumn

\section{Stylistic Attributes}
\label{app:atts-list}
\label{sec:attributes-vals}
\begin{figure*}[h]
    \centering
    \includegraphics[width=0.43\textwidth]{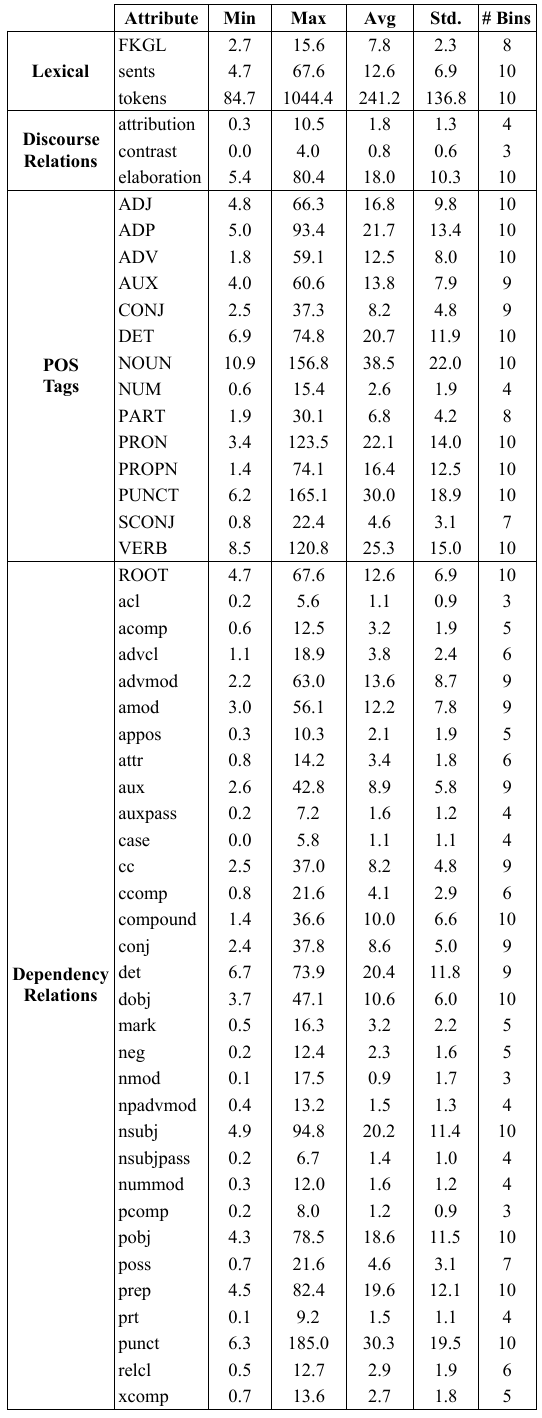}
    \captionof{table}{
    List of the linguistic attributes we model. The minimum, maximum, average, standard deviation and the \# of bins are based on the training data. 
    }
\label{tab:atts-vals}
\end{figure*}

\onecolumn
\section{Stylistic Attributes Performance}
\label{app:attributes-results}
\begin{figure*}[h]
    \centering
    \includegraphics[width=0.55\textwidth]{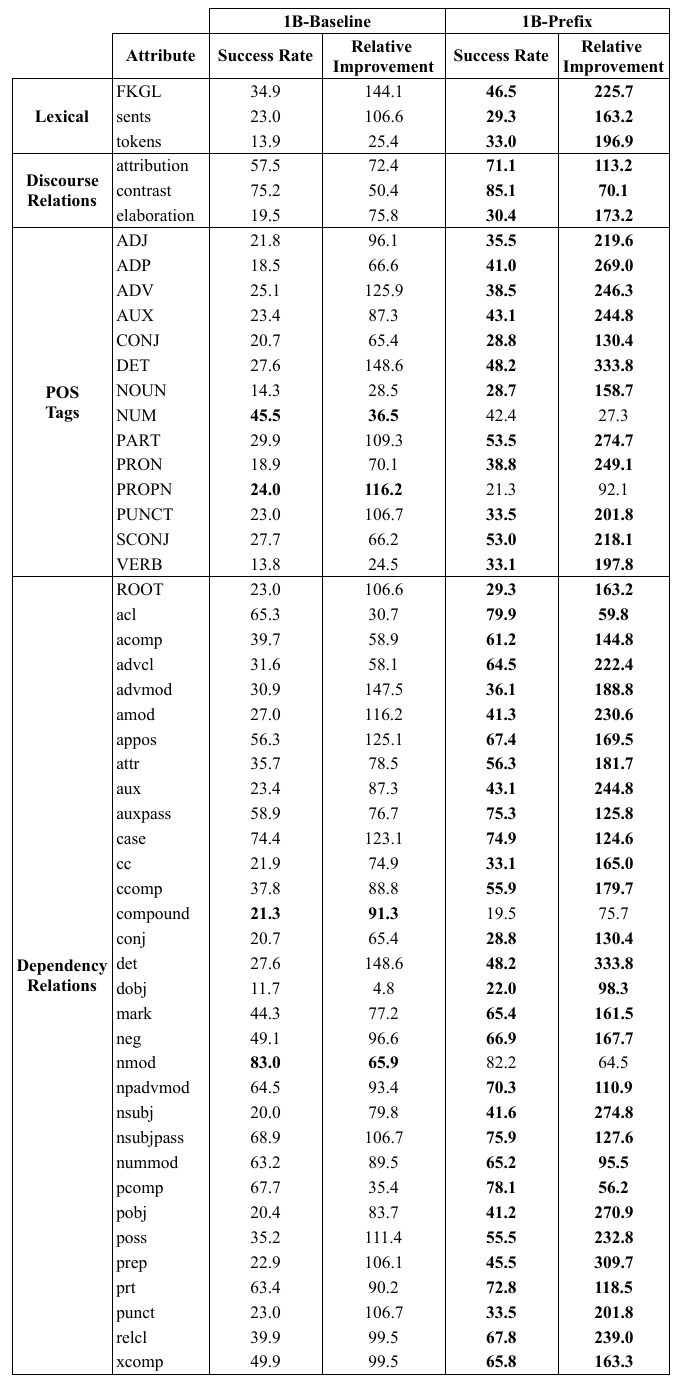}
    \captionof{table}{Attribute-specific performance of the 1B Pythia baseline and Prefix models. The best results are in bold.
    }
\label{tab:atts-results}
\end{figure*}

\end{document}